\documentclass{article}
\usepackage{spconf,amsmath,graphicx}

\title{Siamese LSTM based Fiber Structural Similarity Network (FS2Net) for Rotation Invariant Brain Tractography Segmentation}
\address{}
 \name{ Shreyas Malakarjun Patil$^{\star}$, Aditya Nigam $^{\dagger}$, Arnav Bhavsar $^{\dagger}$, Chiranjoy Chattopadhyay $^{\star}$}

\address{$^{\star}$ Indian Institute of Technology Jodhpur, India \\
    $^{\dagger}$ Indian Institute of Technology Mandi, India}

\begin{document}
%
\maketitle
\begin{abstract}
In this paper, we propose a novel deep learning architecture combining stacked Bi-directional LSTM and LSTMs with the Siamese network architecture for segmentation of brain fibers, obtained from tractography data, into anatomically meaningful clusters.
The proposed network learns the structural difference between fibers of different classes, which enables it to classify fibers with high accuracy.
Importantly, capturing such deep inter and intra class structural relationship also ensures that the segmentation is robust to  relative rotation among test and training data, hence can be used with unregistered data. Our extensive experimentation over order of hundred-thousands of fibers show that the proposed model achieves state-of-the-art results, 
even in cases of large relative rotations between test and training data. 
\end{abstract}
\section{Introduction}
\label{sec:intro}
Inferences about brain structure is 
crucial 
in the diagnosis of numerous disorders and for surgical preparations. A very large number of neuronal connections exist in the human brain that connect various subdivisions, 
which help in communicating between several parts of brain. 
The resulting sequences (or pathways) form what we know as fiber tracts. Tractography on diffusion tensor imaging (DTI) data 
helps in extracting such fiber tracts of the human brain. 
\begin{figure}[htp]
\begin{center}
\includegraphics[width=0.75\linewidth,height=0.418\linewidth]{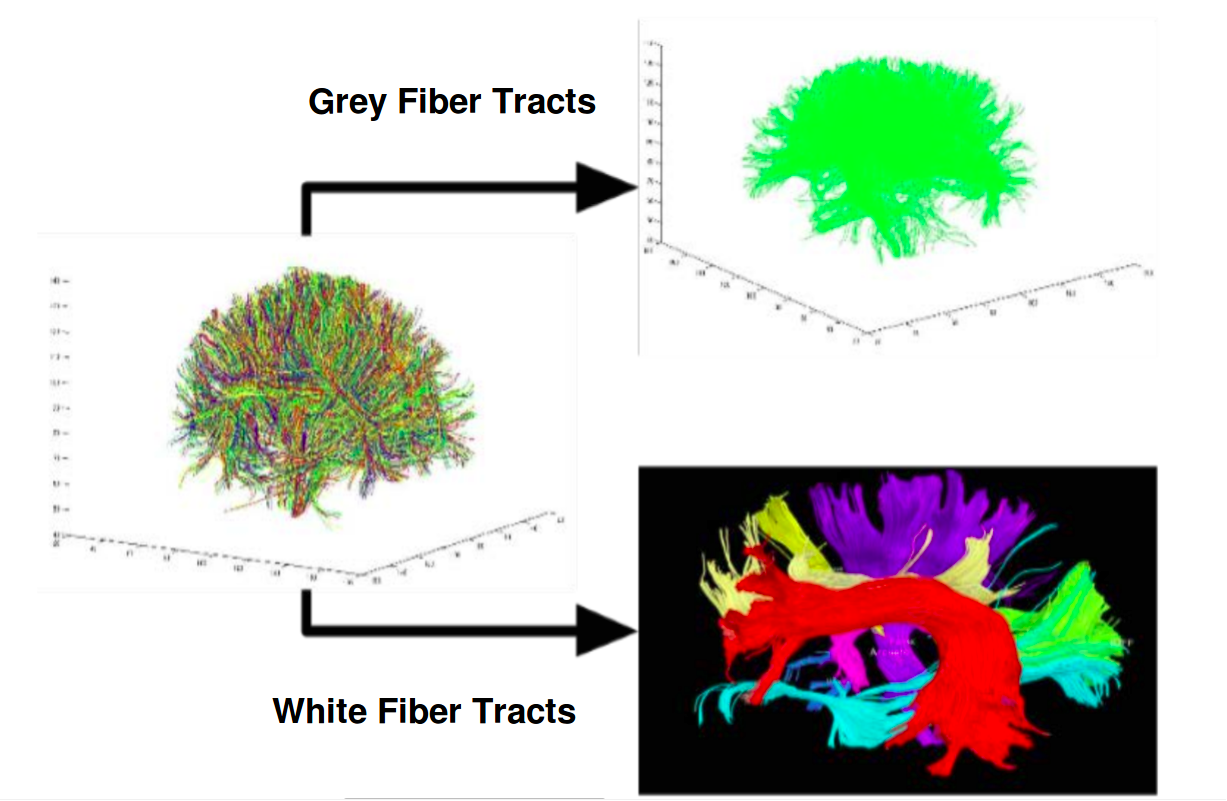}
\end{center}
\caption{Brain fiber tracts}
\label{fig:3}
\vspace{-0.3cm}
\end{figure}

The left portion of Figure \ref{fig:3} shows one such visualization of brain fiber tracts. Estimating clusters of fibers with similar property or 
which follow similar paths (right parts of Figure \ref{fig:3}), can provide very useful information  for high-level inference. 
The human brain consist of millions of such fibers and are hierarchically classified. 
Broadly, there are two kinds of fibers namely gray matter and white matter fibers. The white fibers can be further divided into $8$ important subdivisions: Arcute, Cingulum, Corticospinal, Forceps Major, Fornix, Inferior Occipitofrontal Fasciculus, Superior Longitudinal Fasciculus and Uncinate. As shown in Figure \ref{fig:3} right, 
the grey fibers are spread in the periphery of the brain, and the eight classes of white matter fibers follow distinctive but highly non-linear paths each linking two different brain regions.

Stratification of brain fibers has tremendous amounts of applications in the field of surgical planning 
involving clipping of brain tissues, and for analysis and treatment of degenerative disorders. 
However, manual classification of brain fiber tracts can be tedious as there exist millions of such neuronal connections in any given human brain and requires very high level of proficiency.
Hence, the need for an automated brain fiber segmentation techniques is vital. Moreover, an automated approach should also consider the common issue of non-registered training and testing brain data.  


\textbf{Related Work:} Brain fiber classification has been considered in some contemporary works with both unsupervised and supervised techniques. 
In ~\cite{catani2002virtual}, ~\cite{maddah2005automated} unsupervised approaches 
have been reported. 
Such unsupervised clustering methods involve manual extraction of region of interests.
The authors in ~\cite{o2007automatic} have employed Hausdorff distance as a similarity measure, 
and generate a white matter atlas with spectral clustering. 
The authors in~\cite{5596562} have considered a key set containing only important fiber points of each class and depending upon the proximity of a fiber with a particular set, the fiber is classified either belonging to same class or not. 
In~\cite{wang2011tractography}, a hierarchical Dirichlet process is used to determine the number of clusters. Another supervised approach presented in~\cite{patel2016automated}, selects few major points having maximum curvature and uses these in a clustering algorithm. The training cluster centers are then used to classify test fibers. 
Few approaches based on deep learning employ recurrent neural networks (RNN) \cite{poulin2017learn}, and long-short term memory (LSTM) \cite{1710.05158}.
The approach in \cite{1710.05158} which uses curvature based pruning and bi-directional LSTMs achieves state-of-the-art performance. However, an important concern not addressed in 
recent works has been robustness against handling unregistered training and test data, commonly consisting of a relative rotation. 

\textbf{Contributions:} 1) A new fiber structural similarity network (FS2Net) has been proposed using a moderately deep network built with LSTMs and bi-directional LSTM, in a Siamese architecture, for classification of DTI fiber tracts. 2) The classification is carried out at coarser level ($Grey, White$ matter), as well as on a finer level of white fibers (into 8 classes). 
3) In addition to registered brain data, we also discuss and demonstrate the effectiveness of the proposed approach  even in cases of relative rotation between training and test data. 
4) The proposed architecture and training strategy is computationally efficient; we demonstrate state-of-the-art results with only $11,000$ brain fiber pairs used for training.

\begin{figure}[htp]
\begin{center}
\includegraphics[width=1.055\linewidth,height=1.0518\linewidth]{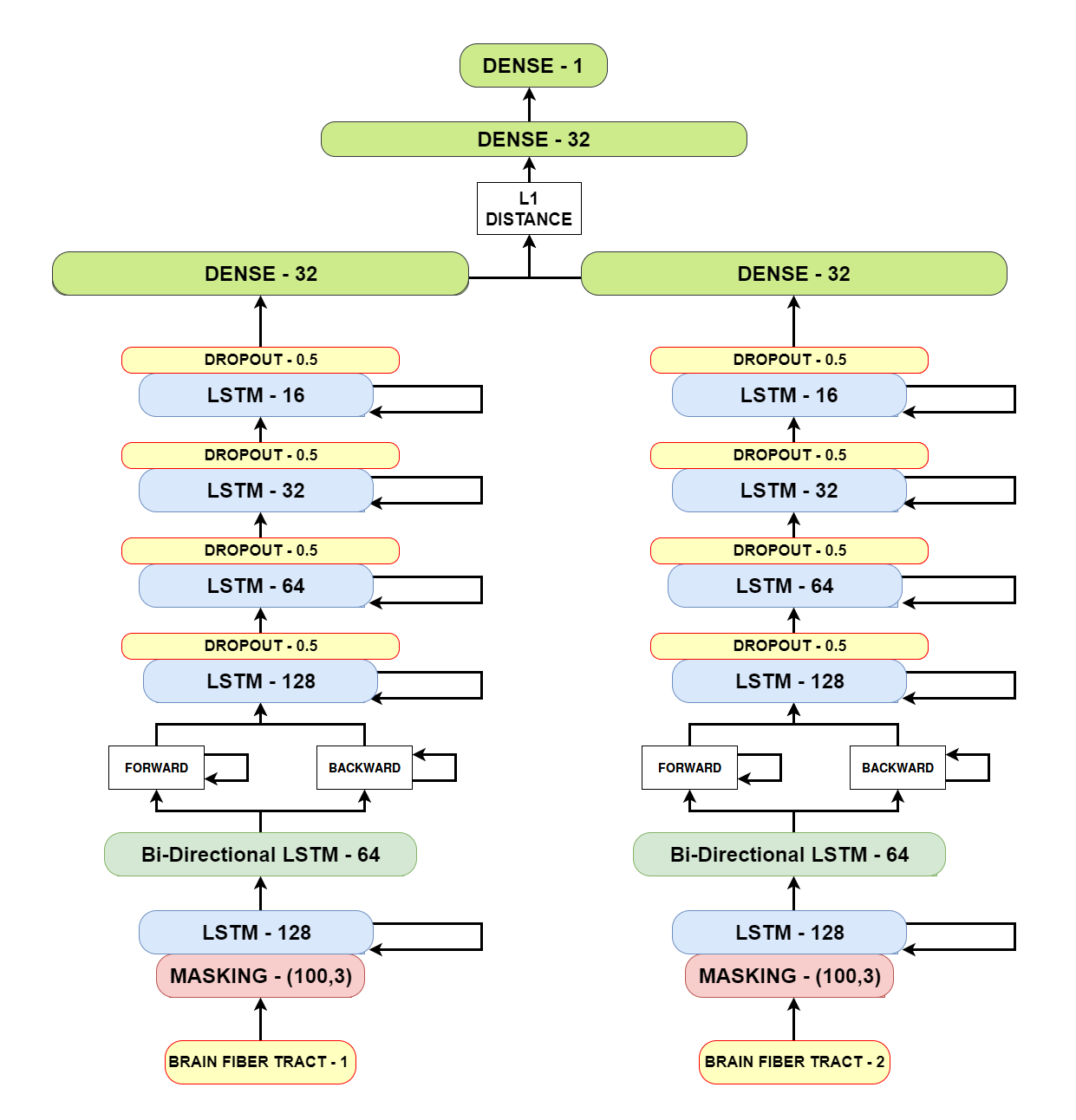}
\end{center} 
\caption{Proposed FS2Net architecture}
\label{fig:1}
\vspace{-0.5cm}
\end{figure}

\section{Proposed Model}
\subsection{Tractography Data}
The tractography data used in this work is acquired from University of Pittsburg, and was used in the Pittsburg Brain Competition on Brain Connectivity conducted with the IEEE Int. Conf. on Data Mining, 2009. The data consists of brain fiber tracts  for three subjects and their respective labels. The ground-truth labels are assigned manually by clinical neurology experts. 
Overall, the data consists of $250,000$ fibers per subject. Each data point of a fiber is denoted by a 3-dimensional vector. The  fibers are varying length containing $36$ to $120$ 3D vectors each. The data is highly skewed in nature, so that almost $90\%$ of fibers belong to Grey matter class and the rest belong to finer fiber classes of White matter. Note that while the data belongs to only 3 subjects, the number of fiber samples to be classified is very large (order of hundred thousands). Also, as discussed later, the proposed network processes pairs of fibers to learn their similarity. 
Thus, the total number of pairs involved are $3 \times 250,000 \times 249,999 / 2$, which, from a pattern classification perspective, is a significantly large data. Interestingly, we obtain state-of-the-art results using only $11,000$ pairs for training. 
\subsection{Curvature Based Data Pre-Processing}
Considering the fiber data is varying length, data pruning has been performed at the beginning to make all inputs of the same size. The pre-processing step 
considers the important curvature points on the fiber. From 
the fiber projection on three planes, the sum of gradients have been computed at the points high curvature. The gradients are computed using one preceding point and one succeeding point, as well as with respect to point four preceding and succeeding points to attain scale invariance. Finally, $25\%$ of the 3D vectors with least gradients are pruned, and the remaining $75\%$ are padded with zeros, if required, to construct features of length $100\times3$. 

\subsection{Model Architecture}
Below, we briefly discuss the components used in the proposed method viz. stacked LSTM, Bi-directional LSTMs (BLSTM), and Siamese network, followed by description of our model, and its training and testing.  



\textbf{Long Short Term Memory (LSTM) elements:} RNNs \cite{sutskever2013training} are a class of neural networks that contain a loop so as to cater for the input from previous hidden layers. These loops act as memory elements 
This allows it to exhibit dynamic temporal behaviour. LSTMs \cite{Hochreiter:1997:LSM:1246443.1246450} are a class of recurrent neural networks that have the additional ability to selectively learn from nodes of previous iterations or what part of the history to retain. 
Any given LSTM block contains three different gates that control the forward information flow namely: the forget gate, input gate and the output gate. The input gate is responsible for the filtering of the input from the previous state, i.e. it learns what nodes of the previous hidden layer to forget and which of these to pass further. Next, the input gate basically decides which values are to be present in the current state, 
and lastly the output gate computes the output of the present or the current state.

\textbf{Bi-Direction LSTMs:} The basic idea behind Bi-directional LSTMs 
is that the output is connected both to the preceding state as well as the succeeding state wherein it is possible to predict based on the information from the past as well as from the future. The neuron of a regular LSTM layer is split into two directional neurons , one for positive time direction(forward states), and another for negative time direction(backward states). By using two time directions, input information from the past and future of the current time frame can be used unlike that in standard LSTM. 

\textbf{Siamese Network based Fiber Comparator:} In Siamese structure \cite{koch2015siamese}, 
two identical networks work in parallel with two different data point inputs, and the dissimilarity is learned between two classes by computing the L1 distance between the corresponding output feature vectors.
The network yields a single binary output 
which specifies whether the two data points belong to the same class (True) or not 
(False). Hence, the network learns the difference between data points from two dissimilar classes or similarities between those from the same class. We have devised a Siamese LSTM based fiber comparator that can capture the deep inter class differences and intra class structural similarity of fibers.

\textbf{Model Justification:} The motivation for selecting LSTMs as the basic nodes for building the proposed network model is due to the reasoning that 
segmentation involves learning the sequencing of the 3D vectors (fiber representations). 
The BLSTMs cater to the consideration that curvature points of a fiber sequence are crucial in identifying its class label, and the curvature is defined by derivatives in both the directions.
The Siamese network architecture has been used to learn the similarities or dissimilarities between paths of fiber tracts for different classes. Importantly, in effect, the network is not just learning the overall fiber trajectory but also the similarities (or differences) between fibers. This aspect allows the model to classify (rotationally) unregistered test fibers relative to the training data, as elaborated in section 2.5. 

\subsection{Training} 
We now discuss the training process for the proposed $FS2Net$. 
Here, 4 models are trained on the tractography data from three subjects: Three separate models from 
data of each of the three subjects, and a fourth model using the combined dataset. The models are trained 
both for coarse and fine levels.

\textbf{Fiber Pairing: } The first step of the training process is the pairing of fibers so as to pass them through the network and consequently learn similarities or dissimilarities. The ratio of number of similar pairs to the number of dissimilar pairs used was $4:7$ for fine level classification and $5:6$ for coarse level classification. The batch size of 11 pairs has been used in our approach, where, in a batch, all the similar pairs belonged to one class, and the dissimilar pair were made corresponding to the same class with equal number of pairs to each of the other classes. The labels for the pairs of same class is given as $1$ (True) and different class as $0$ (False). One of the advantages of pairing and measuring the similarity is that the skewed nature of the dataset is eliminated, as for pairs there must be equal number of all classes of fibers. 

\textbf{Training Process:} After obtaining the pairs from the above mentioned process each fiber pair is passed through the network to obtain a output feature vector of size $32$ after the dense layer as shown in Figure~\ref{fig:1}. The L1 distance is then computed between two feature vectors and finally a single node at the end represents the similarity between the input fibers as $1$  if they belong to the same class and $0$ otherwise. The $ReLu$ activation function is used for all the dense layers except for the last were $sigmoid$ activation is used. 

\textbf{Network Hyper-parameters:} Number of Iteration - $1000$ (In one iteration only a single batch is passed), Batch Size - $11$, Activation Functions - $ReLu$ and $Sigmoid$, Optimizer - $Adam$, Loss Function - $Mean$ $Squared$ $Error$.

\begin{table*}[!t]
\small
\begin{center}
\begin{tabular}{|c|c|c|c||c|c|c||c|c|c|}
\hline

 \textbf{Patient} & \multicolumn{6}{|c|}{\textbf{Accuracy}($\%$)}  & \multicolumn{3}{|c|}{\textbf{Recall}($\%$)} \\ 
 \hline
  &\multicolumn{3}{|c||}{\textbf{Macro}}&\multicolumn{3}{|c||}{\textbf{Micro}} &\multicolumn{3}{|c|}{\textbf{Macro}}\\  
 \hline
 \multicolumn{10}{|c|}{\textbf{Intra: Trained and tested over same brain}} \\
 \hline
 &ANN&BrainSegNet& FS2Net & ANN&BrainSegNet& FS2Net  &  ANN& BrainSegNet& FS2Net \\
\hline
 \textbf{B1} &98.02 &98.88 & 99.23&97.34& 98.12& 99.01&82.1& 94.98 &96.14\\
 \hline
 \textbf{B2} &96.14& 97.84& 98.87 &93.49& 99.96& 99.92&65.2& 80.68&90.34 \\
 \hline
 \textbf{B3} &96.69& 96.42&98.68 &95.96& 97.45&98.97 &57.6& 73.45&81.49\\
 \hline
 \multicolumn{10}{|c|}{\textbf{Inter: Trained on B1 and tested on B2 and B3}} \\
\hline
 \textbf{B2} & 91.32 &---&97.45 & 88.05 &---& 94.58&30.9&---&58.55 \\
 \hline
\textbf{B2} & 94.03 &---&98.98 & 93.66 &---& 95.76&34.4&---&60.56 \\
 \hline
 \multicolumn{10}{|c|}{\textbf{Inter: Trained on B2 and tested on B1 and B3}} \\
\hline
 \textbf{B1} & --- &93.30& 95.51& --- &99.55& 99.67&---&49.00&62.54 \\
 \hline
 \textbf{B3} & --- &94.47&96.11 &---&96.18& 97.89& --- &49.60 &59.09\\
 \hline

 \multicolumn{10}{|c|}{\textbf{Merged: Trained and tested over merged brain data}} \\
\hline
 &94.87&96.65&98.12&93.95& 95.51&97.76 &80.6& 83.46& 90.55\\
 \hline
\end{tabular}
\end{center}
\caption{Performance analysis of the proposed $FS2Net$ for registered data.} 
\label{tab:1}
\end{table*}

\subsection{Testing Strategy}
The above discussion indicates that the network essentially decides whether the two input fiber data belong to the same class or not. This, by itself, is not enough to perform an overall labeling. 
Thus, a default set of labeled brain fiber data is constructed for the purpose of comparison of the test fiber. The test fiber is paired with each one of the default set fibers and passed through the network and then the labeled brain fiber for which this pass obtains the maximum score in a scale from $0$ to $1$ is assigned that class or label. 

In the default set of fibers the rotated fibers of all the classes also have been included so as to give rotation invariance during testing to cater for the unregistered data. Note that rotated brain data is not necessarily required for training the network. As the network is learning only similarity between fibers, it is enough to include the labeled rotated data only in this default set.

\section{Experimental Analysis} 
For each subject, the brain data contains 250,000 fibers. 
As mentioned earlier, we perform a two level hierarchical classification: 
a) Coarse (Macro) Level - In which the fibers undergo binary classification with respect to two classes viz. Grey and the white matter (where data imbalance is a major challenge), b) Finer (Micro) Level - In which the fibers are classified into one of the 8 sub-classes of white fibers. 

The training and testing of our proposed network $FS2Net$ has been performed under the following protocols: 

a) \textbf{Intra Brain Testing:} Partitioning the data from the same patient into training and testing. We have reported results for data from 
one of the three patients.

b) \textbf{Inter Brain Testing:} Training over one of the three patient’s data, and testing on a fraction of the data from other two patients. We have reported our results only when trained over patient 2 (i.e. Brain 2) data and tested over remaining Brain 1 and Brain 3 data.

c) \textbf{Merged Brain Testing:} Training over merged data of all the three patients, such that (randomly chosen) $4000$ of the data points from each brain has been considered for training, while remaining data is used for testing.

All the above mentioned training protocols have been performed for both macro and micro level testing and also for various degrees of rotations.
$Accuracy$ and $Recall$, defined below, have been used to quantify the performance.

\

$Accuracy=\frac{\# correctly\ classified\ fibers}{Total fibers}$

$Recall=\frac{\#\ white\ fibers\ predicted}{Total\ white\ fibers}$
 
\

We have defined $Recall$ as above, so as to quantify the ability of our model to handle skewed data, considering the large difference in white and grey matter fibers numbers.

\textbf{Comparative Analysis:} The results of the proposed approach, in accordance with the testing strategy used in \cite{patel2016automated} and BrainSegNet \cite{1710.05158} for fair comparisons, 
are depicted in Table \ref{tab:1}. 
One can observe that we achieve state-of-the-art results in almost all the test cases. 
The accuracy in the inter-brain case falls compared to that of the intra-brain, due to possible variations between the structure of two brains in terms of size and fiber shape / path. However, the drop in accuracy is much lower that that in the other methods. The recall values in Table 1, signify the superiority of the proposed $FS2Net$ in classifying white fibers (fewer in number), 
suggesting that the proposed model handles data imbalance better. 
The computational efficiency of the proposed network also surpasses that of the previous state-of-the-art. In \cite{1710.05158}, $80,000$ fibers and $25$ epochs have been used whereas, here we have only used a total of $11,000$ pairs in a batch size of $11$ with $1000$ iterations.

\begin{table}[!h]
\small
\begin{center}

\begin{tabular}{|c||c|c|c|c|}
\hline

 \textbf{Patient}  & \multicolumn{4}{|c|}{\textbf{Accuracy}($\%$) - Micro}   \\ 
  \hline
  \textbf{Method}&BrainSegNet&\multicolumn{3}{|c|}{FS2Net}\\ 
  \hline
  \textbf{Rotation Angle}&30&10&20&30\\ 
 \hline
 \multicolumn{5}{|c|}{\textbf{Intra: Trained over same brain}} \\
 \hline
\textbf{Brain 1}&56.68 & 90.21 & 87.74 & 85.56 \\
  \hline
\textbf{Brain 2}&49.51 &91.12 &88.67 &86.23 \\
  \hline
\textbf{Brain 3}&51.32 & 88.96 & 86.78 &83.12  \\
  \hline

 \multicolumn{5}{|c|}{\textbf{Merged: Trained over merged brain data}} \\
\hline
& 43.56&87.91 &86.01 &82.87  \\
 \hline
 &\multicolumn{4}{|c|}{\textbf{Accuracy}($\%$) - Macro}   \\ 
  \hline
   \multicolumn{5}{|c|}{\textbf{Intra: Trained over same brain}} \\
 \hline
\textbf{Brain 1}&58.98 & 91.73 & 88.44 & 84.19 \\
  \hline
\textbf{Brain 2}&52.01 &92.91 &89.99 &88.03 \\
  \hline
\textbf{Brain 3}&53.69 & 90.12 & 89.11 &85.09  \\
  \hline

 \multicolumn{5}{|c|}{\textbf{Merged: Trained over merged brain data}} \\
\hline
& 46.11&88.36 &85.91 &83.38  \\
 \hline
\end{tabular}
\end{center}
\caption{Performance with rotated unregistered test data.} 
\label{tab:2}
\vspace{-0.5cm}
\end{table} 

\textbf{Rotational Invariance:} The results corresponding to the rotation of test cases relative to the training set is shown in Table \ref{tab:2}. Even here, the proposed approach achieves high classification accuracy. Note that as the rotation increases, the accuracy drops slightly, but this may also depend on the labeled samples that are included in the default fiber set used during testing. One can observe that the BrainSegNet method \cite{1710.05158} performs poorly in case of rotations. This is because, it learns the absolute fiber structure / path, which can change significantly under rotation. However, the proposed network achieves rotation invariance via learning similarities rather than absolute structure. 
When a rotated test fiber is paired with a rotated default set fiber of the same class, the network provides a feature vector based on the similarity between these fibers. 
Thus, providing a wide range of fiber rotations in the default set 
enables rotational invariance. Considering a large and diverse dataset with small training samples, we believe that the achieved results are very encouraging. 

\section{Conclusion}
We propose a novel Siamese LSTM-BLSTM 
architecture 
for 
classifying brain fiber tracts, which also caters for unregistered brain data involving rotations. We suggest a two-level hierarchical classification a) White vs Grey matter and b) White matter clusters. Our experimental evaluation shows that the proposed network achieves state-of-the-art classification accuracies, for registered as well as unregistered cases, with significant improvements in some evaluation scenarios. 
\bibliographystyle{IEEEbib}
\bibliography{strings,refs}

\end{document}